\documentclass[lettersize,journal]{IEEEtran}
\usepackage{amsmath,amsfonts}
\usepackage{algorithmic}
\usepackage{algorithm}
\usepackage{array}
\usepackage[caption=false,font=normalsize,labelfont=sf,textfont=sf]{subfig}
\usepackage{textcomp}
\usepackage{stfloats}
\usepackage[hyphens]{url}
\usepackage{verbatim}
\usepackage{graphicx}
\usepackage{cite}
\usepackage{graphicx}
\usepackage{lscape}
\usepackage{comment}

\begin{document}

\fbox{\begin{minipage}{0.9\textwidth}
This work has been submitted to the IEEE for possible publication. Copyright may be transferred without notice, after which this version may no longer be accessible.
\end{minipage}}

\newpage

\title{Developing a Taxonomy of Elements Adversarial to Autonomous Vehicles}

\author{Mohammadali Saffary\thanks{M. Saffary is with the Department of Computer Science and Engineering at Michigan State University, Michigan, United States}, Nishan Inampudi\thanks{N. Inampudi is with the Detroit Country Day High School, Michigan, United States}, Joshua E. Siegel \IEEEmembership{Member, IEEE}\thanks{J. Siegel is with the Department of Computer Science and Engineering at Michigan State University, Michigan, United States}}


\maketitle

\begin{abstract}
As highly automated vehicles reach higher deployment rates, they find themselves in increasingly dangerous situations. Knowing that the consequence of a crash is significant for the health of occupants, bystanders, and properties, as well as to the viability of autonomy and adjacent businesses, we must search for more efficacious ways to comprehensively and reliably train autonomous vehicles to better navigate the complex scenarios with which they struggle. We therefore introduce a taxonomy of potentially adversarial elements that may contribute to poor performance or system failures as a means of identifying and elucidating lesser-seen risks. This taxonomy may be used to characterize failures of automation, as well as to support simulation and real-world training efforts by providing a more comprehensive classification system for events resulting in disengagement, collision, or other negative consequences. This taxonomy is created from and tested against real collision events to ensure comprehensive coverage with minimal class overlap and few omissions. It is intended to be used both for the identification of harm-contributing adversarial events and in the generation thereof (to create extreme edge- and corner-case scenarios) in training procedures.
\end{abstract}


\section{Introduction}
\IEEEPARstart{D}{riving} is inherently dangerous, with the potential for a variety of incidents to cause delays, damage, or harm. Technical solutions have been proposed to address this risk, including lane keep assist, adaptive cruise control, and automated emergency braking, which are elements of broader Advanced Driver Assistance Systems (ADAS). Self-driving, an extreme manifestation of ADAS's constituent capabilities, is widely seen as a solution to minimizing the potential risk of harm. Automation systems and self-driving cars hold promise to reduce the frequency and severity of driving-related incidents\cite{othman2022exploring}.

Self-driving is measured in six levels, with automation systems' potential fully realized at SAE Level 5 (L5) autonomy\cite{sae}. In L5 autonomy, a vehicle drives itself fully in all circumstances and environments, without human oversight or intervention. These systems fail operational or fail safe in the event of a system failure, and have a broad operational design domain (ODD) comprising diverse environments, use cases, and ambient contexts. Self-driving systems of all levels are trained and tested to ensure robust performance on countless driving events, lest previously unencountered scenarios force a dangerous hand-off to a human driver, cause an automation system to perform unexpectedly, or otherwise directly result in negative consequences. While increased exposure to diverse scenarios is essential to the development of safe automated driving systems, some events happen rarely while driving or may be difficult to conceive due to their improbability or infrequency; such edge or corner cases often occur when there is an unusual variation in the parameters of an otherwise typical driving scenario, and are rarely trained against. 

Capturing data from real-world driving is a slow and laborious process with inherent biases against the inclusion of select low-probability, high-impact events. Further, public testing requires a significant initial and ongoing investment, while also running the risk of public endangerment (see Cruise's October 2023 accident\cite{cruiseAccident}). Conversely, simulation is an inexpensive, fast, flexible, and permutable method of testing and developing vehicle hardware and software, including key perception, planning~\cite{karur2021survey}, and control algorithms\cite{Huang_Simulation}\cite{Rosique_Simulation}. Simulation may be used to selectively reproduce adversarial events for data generation~\cite{9991836} and frequently serves as the basis for scenario and data augmentation to increase training data volume and variety.

To generate rich and diverse data, it is helpful to first categorize adversarial events, e.g. via a taxonomy. This allows for training events' distillation into key elements and effective communication and re-creation. Using such a taxonomy can help to characterize common and unlikely failure modes, and be used as the basis for benchmarking self-driving vehicles; also simplifies and allows for the validation of a vehicle's capabilities in dealing with both common and infrequent adverse operating conditions. Such a taxonomy might also be ``reversed'' as a means of increasing insight into the ``unknown unknowns'' (or, lesser-known unknowns, depending on one's beliefs and search methodology\cite{ellsaesser_unknown, Talbot_unknown}) with which automated driving systems typically struggle. Successfully recognizing and responding to a broader range of situations can showcase potential advances in the understanding of the vehicle of its environment and improve the stability and appropriateness of the response. From a comprehensive taxonomy, events might be constructed for the purpose of generating plausible - if improbable - training data.  

We propose creating a taxonomy of potentially adversarial events affecting automated vehicles to help in organizing adversarial events and providing a consistent framework for understanding, discussion, and synthesis. This would simplify categorizing and imagining complex scenarios involving one or more adversarial events occurring in parallel or series, characterizing and creating unique combinations of events that might be unexpectedly encountered. Breaking down possible causes of harm in, to, or by an operating vehicle, and finding commonality among these elements to develop a hierarchy of trunks, branches, and nodes representing the elements contributing towards vehicle safety incidents is an important step towards better representing adversarial events surrounding the operation of self-driving systems and recognizing common root causes for ADAS and self-driving failures.

In this manuscript, we develop and present such a taxonomy, dividing elements into three categories and 15 independent classes. The taxonomy is presented and a detailed description is provided for each branch and node. We then examine the use cases for this and related taxonomies in classifying and categorizing crashes and the environment. Finally, we explore the implications of this taxonomy in easing the development and deployment of future autonomous vehicles and evaluate the taxonomy's efficacy in classifying adverse events.

\section{Literature Review}

To assess and improve the safety of AVs, developers must build confidence that their systems operate correctly in diverse traffic conditions. To that end, cultivating an understanding of the existing challenges in the environment becomes crucial. This section reviews existing approaches in systematic scenario generation and categorization.

Categorization of uncertain and edge-case scenarios is often completed with grouping based on the incident class itself, or by contributing factors. After analyzing real-world accidents, \cite{Alambeigi} finds five event types that describe autonomous vehicle collisions: transition of control, right turn lane at intersections, through lane at an intersection, oncoming traffic at intersections, and sideswiping accidents. \cite{Capallera} creates a taxonomy of events stemming from distinguished uncertainties and limitations recognized by vehicle manufacturers as published in their owner manuals; this resulted in 26 micro-categories fitted into 6 macro-categories with potential for duplicates, in which an event may be included in multiple macro-categories. \cite{Ramirez} also creates an event classification system at run-time, focusing primarily on sensor uncertainties.

Events may be further decomposed into their contributing elements. \cite{Bagschik} proposes a five-layer model for representing driving scenes. Layers include road geometry, traffic infrastructure, temporary modifications to the previous two layers, objects, and the environment. \cite{Sauerbier} extends this model with digital information as a sixth layer. \cite{Liu} and \cite{Halilaj} each present an element classification model respectively for autonomous vehicle crash scenes and for representing driving dimensions in knowledge graphs. 

\cite{Bagschik} and \cite{Liu} develop models in which scenario-descriptor elements may be linked and dependent, unlike models created by \cite{Alambeigi} and \cite{Halilaj}. \cite{Capallera} mixes the two strategies to have the macro-categories be inclusive, but the micro-categories be non-inclusive. 

Vehicle scenario generation and testing are divided into data-driven derived from real-world accidents and knowledge-based where scenarios are conceived by experts \cite{Riedmaier} \cite{Ding}. The California DMV AV collision report dataset\cite{CaliforniaAVReport} is one representative source for data-driven event scenario generation \cite{Alambeigi} \cite{Liu}. By using the vehicle owner's manuals, \cite{Capallera} extracted knowledge-based data on advanced driving assistant systems (ADAS) limitations. \cite{Liu} also presents an adversarial scenario generation type in which scenarios are generated with the ego vehicle in mind. 

Long-term, unplanned \cite{Zhong} and model-based \cite{Chao} general testing has been implemented to assess the behavioral features of the vehicles. These tests are best for evaluating the conduct of an autonomous vehicle (like planning, driving style, and cooperativeness \cite{Halilaj}) in large traffic or in ego-specific scenarios \cite{Chao}. As important as these tests are to evaluate behavioral success, behavioral failures should also be studied. Thus short-term scenario-specific tests and events must also be planned and studied physically \cite{Zhong} and in simulation\cite{Chao}. These tests are optimized to find events and environments where autonomous systems might fail to perform \cite{Riedmaier}. Events where vehicles failed to successfully perform then can be turned into valuable datasets like \cite{CaliforniaAVReport} and inform the creation of uncertainty classifiers \cite{Ramirez} or test procedures. 

For developers to build a comprehension of the environment of the AV, they need to fully understand the potential role of each element in contributing to challenges in automation. Purely data-driven models risk being incomplete as they rely strictly on what has happened not what might happen, emphasizing the importance of knowledge-driven models. Studying the literature and identifying potential opportunities for improvement encouraged us to develop a taxonomy with elements that are not interdependent and can be tested against in isolation or concurrently. Our taxonomy is conceived using both example-driven and knowledge-based data to be exhaustive of all possible adversarial elements. This taxonomy creates the opportunity for AV developers to decompose scenarios, understand the role of the involved elements, and test their vehicles against comparable adversarial scenarios in simulation to extract failure conditions.

\section{Methodology}

Addressing the identified needs, we create a data- and knowledge-driven taxonomy that includes and describes comprehensively the most significant elements that contribute to an automation system failure or accident. Such a task is challenging as all of the most commonly known elements must be recognized and categorized. Elements contained in the ``unknown unknown'' set cannot be categorized before being identified and established as contributing elements. To ensure adequate representation, we proposed using data from broadly-curated autonomous vehicle test accident reports to organize a set that encompasses known elements associated with an event, whether those elements fully, partially, or potentially precipitated in the incident. 

The most valuable and desirable accident reports for our purposes are ones that describe the conditions, actions, and intentions of the actors, environment, and equipment involved. Thoughtful pre-crash descriptions can provide appropriate context for a scene whereas normal accident reports are concerned with the location, the damage, and the party at fault at a crash scene. Such a constraint leaves us with few datasets that are consistent in data recording, have a substantial number of samples, and are not demographically biased toward specific AV manufacturers.

After specifying taxonomy objectives and requirements, we identified data against which to design, test, and validate our classification system. We checked for existing crashes on avcrashes.net \cite{avcrashes}, a widely-referenced webpage supported by the Czech Republic Ministry of Transport to create a unified database of autonomous vehicle accident reports. This website provides descriptions of crashes along with data and self-reported statistics on fault, injury, and damage to the driver and vehicle. The website contained 588 records as of December 2022; these were used for sculpting the initial drafts of the taxonomy. At the time of writing this article, there were a total of 725 accidents reported on this site, reflecting broader adoption of automated vehicles. Each accident report is parsed by one of a consistent set of human operators who confirm the details of the event and enter it into a database. Samples come from around the world, with a majority from California USA, which requires manufacturers testing autonomous vehicles in the state to report any collision that resulted in property damage, bodily injury, or death~\cite{CaliforniaAVReport}.

We investigated these samples and built a diverse set of events involving common and unique elements that have a potentially adversarial effect on the normal operation of an AV. Most of the samples were curated from operations within naturalistic driving scenarios. An initial version of this sample set tended to cluster around a few causal factors due to the non-uniform distribution of vehicle collisions and their precipitating events. We therefore drew additional inspiration beyond the avcrashes.net dataset from eye-catching news headlines, personal experiences, and historically memorable vehicular accidents to develop a more exhaustive pool of AV-adversarial samples including infrequent edge and corner cases with potential significance in failures. We also added to our list events that would not be immediately dangerous, but could cause potential harm to the vehicle, its occupants, bystanders, or property if not appropriately remediated, e.g. improper vehicle repair and maintenance. As AV adoption increases, mechanical and other faults may reasonably become increasingly common causes of collision. 

A thorough examination of each element of the factors identified in the dataset and among the attention-grabbing headlines exposed common and unique themes contributing to failures. Variations of each element and multiple circumstances where such elements might appear were considered to ensure the differences between elements were defined. Common elements were put into classes with an appropriate label that best described all included elements of that group. This process was repeated among the classes of elements until no element was left out and all groups were entered into hierarchical form. This resulted in a taxonomy that includes all elements as leaves and their categories as the branch of the taxonomy tree. 

To properly distinguish each of the elements, we follow the template presented in \cite{Ramirez}: each leaf includes a description to provide context for what it communicates, what it does and does not include, and how it is differentiated from other leaves. Where a significant corpus of public documentation exists, we also discuss what mitigation strategies have been proposed or implemented to address the identified factors, combining information from academia, standards, industry, and legislation. Finally, we present examples where such elements have either caused an accident or altered the normally safe environment of a vehicle such that requires a change in driving behavior, for example, navigating around an airplane landing on a highway, or a bicycle falling off the back of a leading vehicle.

During the evaluation, avcrashes.net \cite{avcrashes} became unavailable for a period, so we directly accessed California Department of Motor Vehicles Autonomous Vehicle collision reports \cite{CaliforniaAVReport}. These reports are standardized among reporting parties and include a full description of the event and the environmental features at the scene. We gathered an additional 117 samples, from January to September of 2023, from \cite{CaliforniaAVReport} to test the validity of our taxonomy. These are cases where any of the active autonomous system, ego driver, or other agents can be at fault. We inspect each test case to determine which elements were included in the scenario, what implications the existence of each element had on the accident, and, most importantly, whether the taxonomy includes the identified elements and properly conveys the significance of such elements. Finally, we show the frequency at which each element appears as the fundamental contributing element and the difficulty of placing the element in the taxonomy.
  
\section{Taxonomy for Organizing Adversarial Events}

\begin{figure*}
    \centering
    \includegraphics[angle=90,origin=c,width=\textwidth]{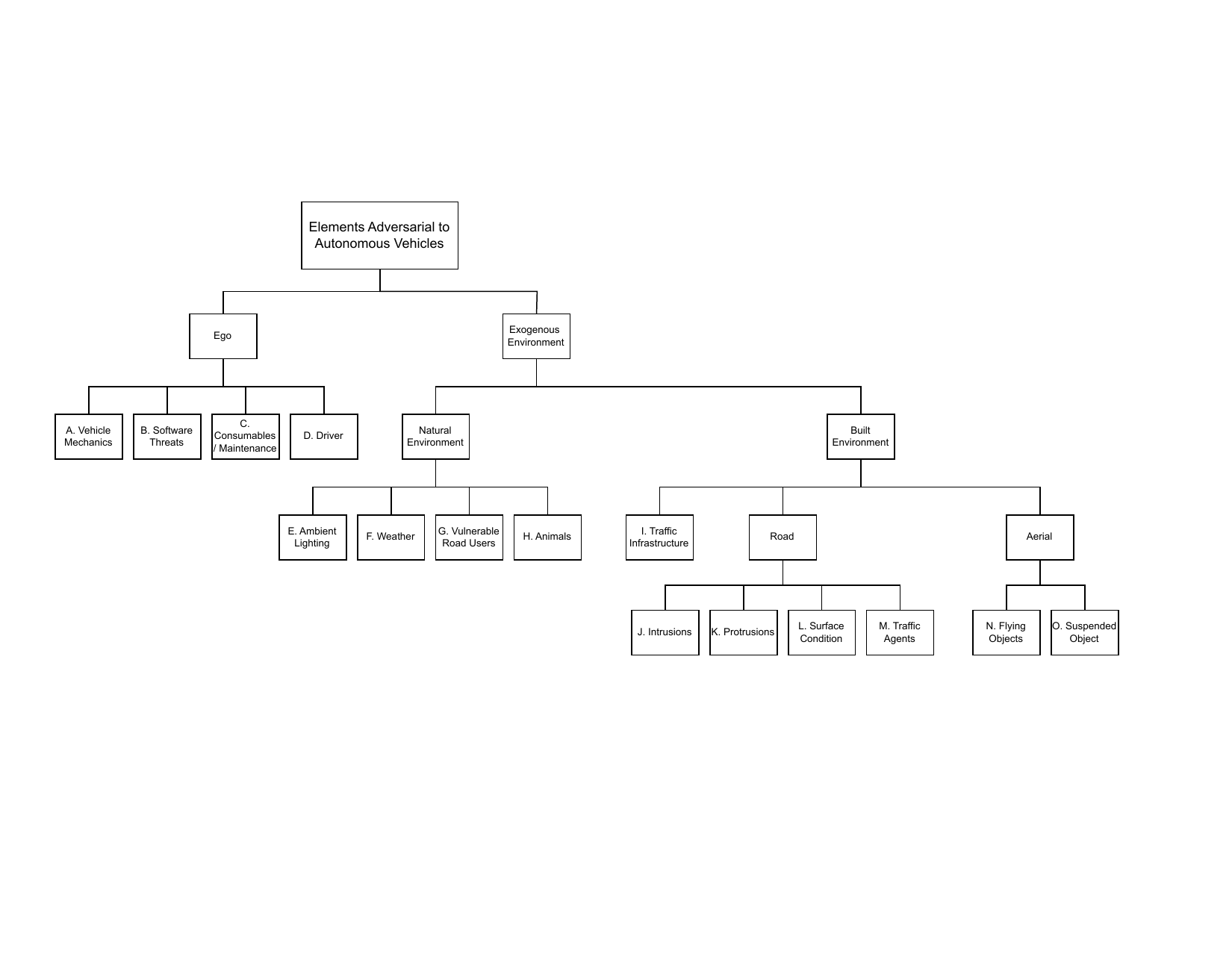}
    \caption{Taxonomy of adversarial elements.}
    \label{fig:tax_full}
\end{figure*}

A taxonomy for organizing adversarial events provides a framework for understanding and classifying or generating consequential and/or informative events by grouping similar events and providing a common vocabulary for exploration and analysis. In this section, we present a taxonomy for organizing adversarial events in the context of automated vehicle development. For each node, we present a definition, a description of the class's scope, and its significance in ensuring the safe operation of autonomous vehicles. If a node is a terminal element (a leaf of the taxonomy), we also present examples and relevant investigations (such as standards, laws, latest academic articles, or promising techniques) that apply to vehicles across autonomy levels. 

The taxonomy aims to include comprehensively any contributing sources of a vehicular accident with respect to the vehicle's environment. The first branching is the separation of the inner context (ego) from the outer (exogenous) environment of the vehicle. 

\subsection{Ego}
The car itself, as well as its driver, are often referred to as \textit{ego}, Latin for ``I,'' or ``self.'' Elements of this category may be seen as having a direct influence over the control of the vehicle. The automation system must not only be responsive to the environment outside the car, but also to any danger occurring as a result of failing components or software, inadequate consumable quantity, condition, or remaining life, and in the case of partially automated driving systems, an occupant's level of attention and state of mind. 

In terms of self-protection, electromechanical and software fail-safe, often with fail-functional provisions present in modern vehicles' safety systems. These techniques often result from functional safety requirements, internal software engineering strategies, and the consistent use of the V-model for development ~\cite{V_model}. However, heightened levels of autonomy require additional focus on unique elements such as hand-off and other transient human/vehicle interactions that extend beyond the purview of typical in-vehicle software. 

Beyond the in-car hardware and software, maintenance and consumables require continual attention (attention that may not be given due consideration in a vehicle with which the occupants are expected to disengage). For many AV systems, these cannot reliably address maintenance issues independently unless additional software is implemented \cite{siegel2016engine, Siegel_obd_oil, siegel2015smartphone, siegel2016smartphone, siegel2017air, siegel2018automotive, siegel2021surveying, terwilliger2022ai, terwilliger2022improving}. Continual sensing and techniques for alerting the driver and owner of the vehicle of impending repair needs become an important avenue to keep the vehicle in proper running condition. Depending on the type of vehicle in operation, e.g. personal versus shared mobility, responsibility for maintenance may fall to several parties, many of whom will not have ready access to observe and inspect the vehicle. 

In Level 3 and below automated driving systems \cite{sae}, the driver is the fail-safe for the autonomous system. If the AV is unsure of how to react to its surroundings, it will hand over controls to the driver. Thus, ensuring the alertness of the driver is important as they are liable for any damages if the AV is disengaged from the vehicle controls \cite{Morando_driver}.

Within the Ego branch, each element has the potential to protect against or cause issues for an automated vehicle. Specific elements are detailed as follows.

\begin{figure}
    \centering
    \includegraphics[width=0.45\textwidth]{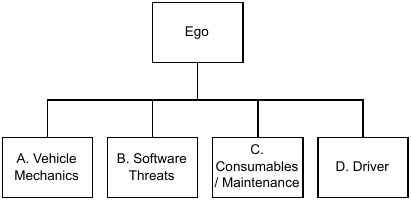}
    \caption{Taxonomy of ego-related elements.}
    \label{fig:tax_ego}
\end{figure}

\begin{itemize}
\item[A.]{Vehicle Mechanics}
\end{itemize}

Mechanical safety measures keep occupants safe and make sure they suffer minimum consequences in case of an accident. These include passive measures like the proper positioning and operation of airbags, tempered and laminated glass, impact-absorbing chassis, head restraints, seat belts, and more. There are also safety measures to help avoid accidents from happening, including lighting, reflective devices, proper seating position and field of view, warning devices and sensors, etc. New vehicles are required to pass certain safety standards in each country before they are allowed to be sold. Many of these safety standards are meant to measure and ensure the operation of a vehicle's mechanical components. In the U.S., these safety regulations are called Federal Motor Vehicle Safety Standards (FMVSS) \cite{FMVSS}. FMVSS sets a minimum standard to which the automakers must adhere for the benefit of the future owners of their products. Organizations like NHTSA~\cite{NHTSA} and IIHS \cite{IIHS} perform tests to ensure new vehicles sold in the US can pass the safety standards. They perform crash tests to ensure the vehicles can withstand impacts and test the efficacy of all safety devices to judge the vehicle's overall safety rating. Safety measures such as roof crush resistance (FMVSS section 216), hood latch system (section 114), and flammability of interior materials (section 302) are examples of mechanical failure mitigation laws that FMVSS mandates. Manufacturers, too, have their own (often secretive) testing regimens. Failure to ensure the mechanical safety measures of the vehicles sold could cause harm to road users and/or result in safety recalls as made apparent by the famous Ford Pinto recall \cite{Pinto_recall}. FMVSS and internal testing provide a means of reducing potential harms during an AV failure event, but may also present new failure modalities contributing to a collision, e.g. a failed element of a lighting system. 

In this class, we also include failures of sensing devices \cite{Campbell_AV_sensors} (cameras, LiDARs, GPS, etc.), onboard computing devices, and electronic interconnects. Some faults cannot be alleviated by a system reboot, update, reinstall, or malware scan and purge. Sensor fusion \cite{Campbell_AV_sensors} is a solution proposed for this problem. One advantage of sensor fusion is that the effects and responsibilities of a malfunctioning sensor can be dispersed to the various other sensors fitted to the vehicle. 

\begin{itemize}
\item[B.]{Software Threats}
\end{itemize}

Vehicle security, before automated vehicles and still, remains a concern. Traditionally, it has been considered as thieves stealing the car or components like the audio system, wheels, or items inside the vehicle. To address these concerns, automakers have proposed and implemented solutions like multiple different keys for vehicles, locking glove boxes, immobilizers, anti-theft alarms, and more. With the advent of more and increasingly connected vehicles \cite{Siegel_its_survey} and increasing software prevalence in vehicles, new security concerns have arisen. \cite{Valasek2018} and \cite{Thing_security_taxonomy} describe the potential of attackers finding access to vehicle ECU through any surface and from any distance, while external hardware can present unique opportunities and threat vectors for extra-vehicular connectivity ~\cite{wilhelm2015cloudthink}.

Software has unique vulnerabilities. Crashes, bugs, and data leaks are among the primary concerns \cite{Wsj_Tesla_software}. Modern cars utilize increasingly digital systems, partially or fully replacing analog devices with the likes of steer-by-wire and brake-by-wire systems, where the driver's inputs are only electronically relayed to the actuating components rather than through mechanical linkage. \cite{Valasek2015} and \cite{Koscher} demonstrate how using CAN messages the vehicle can be manipulated through physical or remote access.

As vehicles become more automated through software integration more attack vectors become available. \cite{Bouchelaghem} \cite{Nanda} \cite{Kim} and \cite{Cui} describe both attack vectors like sensors, CAN Bus, mobile apps, V2X communications and how to address them using firewall, authentications, data-freshness, and more. Design frameworks and architectures~\cite{suo2018merging,falco2020distributed,kent2020assuring} can help to address some, but not all challenges. Simulations can provide an effective platform to enhance detection and defensive measures of vehicle systems \cite{Moukahal}.

\begin{itemize}
\item[C.]{Consumables and Maintenance}
\end{itemize}

Performing maintenance and managing consumables are one responsibility of vehicle ownership and operation. Carrying out operational checks for components such as lights, suspension, glasses, and panels is part of maintenance. On the other hand, consumables are fluids, tires, and other parts that have a shorter life expectancy. Failure to address these measures in any vehicle \cite{Tire_maintenance} will negatively impact its performance, shorten its life, and potentially turn the vehicle into a hazard for any road user. To address these issues, work has been done to identify maintenance and wear considerations to raise operater awareness such that they might keep vehicles in a healthy operating condition and mitigate worst-case failure scenarios.

Vehicle in Use Safety Standard \cite{Inspection} is a guideline presented by the U.S. government against which vehicle equipment should be scrutinized. The US government does not legally require these instructions to be followed, though they are used as a guide for each state to craft its own set of inspection regulations. Many states require the owner to professionally perform a safety inspection on their vehicle every 1 or 2 years. Some states, like New York not only perform safety inspections, but also emission inspections \cite{NY_Inspection}. New York requires safety examinations on seat belts, brakes, steering system, suspension, chassis, tires, lights, all glasses, wipers, mirrors, and the fuel tank.

Some sections of FMVSS \cite{FMVSS} are concerned with the function and monitoring of consumable materials like No. 116 and No. 138 (respectively vehicle brake fluid and tire pressure monitoring systems). Furthermore, modern vehicles are capable of measuring the quantity and quality of fluids \cite{Jun_oil_measure, Siegel_obd_oil} and equipment \cite{Siegel_patent}. Thus, vehicles can recommend services pertaining to such consumables when required.

\begin{itemize}
\item[D.]{Driver}
\end{itemize}

Vehicle occupants include drivers who, in many cases, have studied laws and regulations related to driving. However, awareness of rules and regulations is only part of the battle, particularly as the role of the driver transitions from actively engaged to passive passenger. 

A driver's (lack of) attention has historically been a frequent contributor to road collisions, with distraction causing an average of 14\% of crashes and 8\% of crash fatalities per year \cite{crashstats}. \cite{Bakowski_reactiontime} shows that it takes a minimum of 1 second for drivers to look up from a distraction and at least 2 seconds to start applying brakes in head-on collision situations, with longer times required to stably resume control. Other factors impacting driver attention include drunkenness, drugs, and drowsiness. 

Car companies have been deploying multiple systems to 1) measure and gain the driver's attention \cite{Lee_driveralert} or 2) correct the car's behavior in accordance with its environment \cite{Coelingh_AEB} \cite{Mammeri_laneKeep}. The latter system can act as a solution if the first fails to grab the driver's attention promptly. Two common accident prevention systems are emergency brakes and lane-keeping assistance. The impact of these systems has encouraged the development of tests \cite{Euro_ncap} \cite{IIHS}.

All vehicles, regardless of autonomy status, will be tested against safety standards such as FMVSS. AVs will have to show further capabilities in mechanical safety protection, software threat mitigation, occupant tracking, and consumables monitoring. Ambiguousness about any possible threats in the vehicle brings the safety of the AV into question. Thus safety assurance scenarios can be run in simulations to both train and test a vehicle's capabilities in identifying and managing threats in the ego environment.

\subsection{Exogenous Environment}
The exogenous environment contains every element outside the ego vehicle that can pose a risk to the health and operation of the vehicle. We divide this category into two groups: the natural or built environment. 

\subsubsection{Natural Environment}

Nature, as presented in this taxonomy, consists of any environmental element that is not significantly influenced by the ego vehicle's activity. These elements are marked by their unpredictability such as ambient lighting, weather, animals, and Vulnerable Road Users (VRUs).

Ambient lighting refers to the luminosity of the immediate environment of the vehicle. The most significant contributor to ambient lighting is sky lighting. Depending on the time of the day, the driver may be required, or choose, to turn on the headlights of the vehicle. Although sunset and sunrise are predictable, what adds to the unpredictability of ambient lighting is the weather's involvement. The cloudiness of the sky can affect whether it becomes dark enough during the day to require the headlights to be turned on. Weather forecasting is imperfect, and weather is not controllable. Particulates, heavy snow, and fog can affect visibility to the point where operational considerations must change. 

VRUs and animals are similar as they are both conscious and respond to stimuli. Both may act in unpredictable ways and encroach on a vehicle's operating space. Being able to recognize a vehicle and its trajectory is a potential advantage and a disadvantage. These agents can actively try to avoid vehicles on the road, e.g. by taking advantage of the infrastructure. Conversely, they might be acting maliciously in trying to obstruct the road, and intentionally cause harm to the vehicle, passengers, or themselves. They might also simply be non-cooperative~\cite{seams24paper}. Likewise, animals can cause harm to the vehicle if they are confused or feel threatened.

Below we will discuss in detail the natural elements of the environment and their involvement in the operation of an autonomous vehicle, how they might cause problems, and what countermeasures have been proposed and implemented.

\begin{figure}
    \centering
    \includegraphics[width=0.45\textwidth]{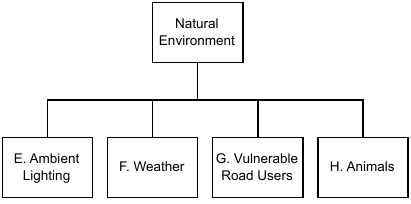}
    \caption{Taxonomy of elements occurring in the natural environment.}
    \label{fig:tax_nature}
\end{figure}

\begin{itemize}
\item[E.]{Ambient Lighting}
\end{itemize}

A favorable period for driving is during the day when light can easily reflect from all objects and reach the driver's eyes or camera's lens. There are some other times when ambient lighting is unfavorable for a driver or certain implementations of a software-enhanced safe driving system, e.g. after sunset and before sunrise. These periods are more common during winters when daylight is shortened. Reduced lighting hinders a human or computer's ability to perceive and avoid potential hazards as exemplified in the Tesla Autopilot crash in the early hours of morning \cite{Banner_crash}.

This is why headlights have been mandated by international regulatory bodies \cite{unece}. In the United States, FMVSS Section 108 \cite{FMVSS} applies to vehicle lighting. NHTSA and IIHS \cite{NHTSA} \cite{IIHS} perform tests on production vehicles to ensure they are compatible with the regulations. Every state may have laws about when the headlights should be turned on and when the other lights should be used.

By way of example, as one approach to addressing limited visibility, the Michigan legislation \cite{Michigan_light} authored in 1949 requires Michigan drivers to turn on vehicle headlights at least 30 minutes after sunset until at least 30 minutes before sunrise, and ``at any other time when there is not sufficient light to render discernible persons and vehicles on the highway at a distance of 500 feet ahead.'' For Washington and California, this distance is 1000 feet \cite{Washington_light} \cite{California_light}. California law also obligates headlights if conditions require the continuous use of windshield wipers such as rain, mist, snow, and fog.

High beams are suggested to be turned on at night during the extreme dark. These lights shine at a higher height and improve the visibility of objects at a longer distance. However, to avoid direct glare for other drivers it should be turned off when any visible vehicle is in front of your vehicle. California Law has set the distance to 500 feet for oncoming vehicles and 300 feet for vehicles in the same direction ~\cite{California_light}.

Daytime running lights are not required in the US despite their mainstream use. FMVSS ~\cite{FMVSS} does not require their presence, but does specify how such lights are to be implemented if used. Some countries require them like UK ~\cite{UK_Daytime_runninglight}. The purpose of DRLs is to grab the attention of oncoming drivers and pedestrians and make the vehicle more visible to other road users no matter what time of day or environmental visibility conditions. A failure in any lighting - in-vehicle or ambient - may be a contributing factor in automated vehicle collisions. 

Through the use of sensor fusion techniques, autonomous vehicles can utilize other perception methods to function, with enhanced performance (in the presence of multiple information sources), or equivalent or reduced performance but the ability to fail-operational or fail-safe. \cite{Jahromi_sensorfailure} shows how LiDARs and Radars can take over the entire perception of the vehicle if the effectiveness of cameras is reduced to the point of failure. Furthermore, autonomous vehicles can use luminescence sensors to turn on headlights and high beams when necessary. It is important for autonomous vehicles to optimize these capabilities before deployment in simulations.

\begin{itemize}
\item[F.]{Weather}
\end{itemize}

Weather refers to atmospheric conditions such as precipitation, wind, humidity, particulates, and temperature. Such conditions although benign towards a vehicle, can have adverse side effects on the safe operation of vehicles. conditions like heavy precipitation, fog, and dust storms can alter the accuracy of perception systems \cite{Zhang_sensorVSweather}. This is made apparent by development autonomous vehicles performing worse when driving in the rain compared to dry track \cite{Hyundai_VSweather}

Determining the weather conditions can be done by taking images of the road and feeding the image as an input to a classification model \cite{Elhoseiny_weather_DL} \cite{Zhang_weather_classification} \cite{Dhananjaya_weatherlight}. It can also be gathered from weather sensors integrated into infrastructure \cite{Wang_weather}, or polled from remote weather stations connected to the internet \cite{Weather_REST_API}.

Cameras are often used alongside LiDAR and RADAR in contemporary automated vehicles \cite{Yurtsever_AVsummary}. They have also become commonplace in human-operated vehicles, e.g., for collision avoidance systems. Using the cameras already equipped on the vehicle can give us locally applicable data on the weather pattern around the vehicle to be analyzed by the vehicle's internal computers or in collaboration with infrastructure processing power \cite{Wang_weather}.

Another way to obtain detailed information on the weather is through external data sources like web APIs ~\cite{Weather_REST_API}. Modern vehicles have Internet connectivity \cite{Siegel_its_survey, siegel2018algorithms}; using a global navigational satellite system unit and an Internet connection, real-time weather can be queried from web sources and databases. 

Meteorologists may do a better job at local forecasting than what a vehicle itself may perceive. However, in some cases using a vehicle's LiDAR has the benefit of having access to hyper-local weather patterns that may change from one street to another \cite{Vargas_Rivero_lidarweather}, while cameras can identify rain, snow, fog, and other contentious operating environments.

Having prior knowledge about the weather can help in predicting road conditions, the clarity of the sky, and traffic. Since hyper-local weather conditions can unexpectedly change, an autonomous vehicle needs to be able to identify and safely deal with any extreme weather condition lest it becomes a hazard to its operation. 

\begin{itemize}
\item[G.]{Vulnerable Road Users}
\end{itemize}
Vulnerable Road Users (VRUs) refer to individuals at an increased risk from, and with a higher susceptibility to, road hazards. VRUs include pedestrians, wheelchair users, and scooter, motorcycle, and bicycle riders~\cite{barnett2020automated}. These entities each have dedicated sections of the infrastructure such as sidewalks, bicycle lanes, or normal roads. Their mobility may or may not allow maneuvering with fewer restrictions than vehicles. Thus, they may appear in areas where they are not usually expected. Their vulnerability also increases the risk of serious injuries; NHTSA statistics for 2021 show VRUs compose 20 percent of accident-related mortality \cite{crashstats_vru}. There are also increased moral and ethical considerations~\cite{siegel2021morals, kassens2021choosing} with pedestrians and other VRUs not present when considering incidents not involving living things, resulting in additional study~\cite{pappas2022game}.

Pedestrian collision accidents have been common since motorized vehicles emerged\cite{Driscolls_death} and still \cite{cruiseAccident}. Traditional solutions for pedestrian avoidance include radar-based forward collision alarms and avoidance systems that notify the driver of imminent frontal crashes and apply brakes automatically \cite{Kelley_patent, Mai_patent}. Contemporary systems could include cameras as well \cite{Iftikhar_pedDetection}.

IIHS ~\cite{IIHS} and Euro NCAP ~\cite{Euro_ncap} perform tests of how well cars can detect VRUs and their braking reaction time. Furthermore, they investigate VRU impact and injury level in case of a collision. This incentivizes manufacturers to build cars and develop systems that increase VRU protections such as active hood systems \cite{Green_patent}.

Autonomous vehicles can endanger VRUs just as much or more than human-driven vehicles as was illustrated in the October 2023 Cruise incident \cite{cruiseAccident}. Success in this process may include detection \cite{Iftikhar_pedDetection}, control \cite{schratter2019pedestrian}, and/or communication \cite{Pappas_pedInteraction}.

\begin{itemize}
\item[H.]{Animals}
\end{itemize}
Hitting any animal at speed is likely to cause damage to the vehicle and injury or loss of life. Wildlife collisions are concentrated around areas with heavy vegetation, as this reduces drivers' ability to perceive and react\cite{Carvalho_vegetation}, or in areas with sparse fencing \cite{Kioko_animalfencing}. Animal collisions may also happen in dense, urban areas as exemplified by the Waymo vehicle that hit a dog \cite{Waymo_vsDog}. 

Infrastructure solutions such as overpasses and tunnels \cite{Nyt_animal} have been implemented at places to minimize collisions, however, animal detection and avoidance remains an important issue. Autonomous vehicles can avoid incidents by using high dimensional image \cite{Saxena_animalDetection} and other forms of signal processing \cite{Nandutu_animalDetection}, at times together \cite{Forslund_animalNight} to detect animals and react accordingly. Animal packs, such as livestock, might also be detected \cite{Meena_animalCounting}, though their behavior may differ from individual animals. 

The recognition of livestock and wild animals, and correct reaction to their existence become increasingly important as driving characteristics become more extreme. Simulations using one, few, or many animals can easily be done to prepare autonomous vehicles against these scenarios.

We see how each sub-category of nature poses its own dangers, how their inherent unpredictability makes them part of the exogenous environment, how they are separated, and how the community is proposing to solve and deal with potential problems. The use of simulations can ease the learning difficulties of sufficiently finding extreme examples of these natural elements and improve the AVs understanding of the natural environment.

\subsubsection{Built Environment}
The built environment is every element in the vehicle's surroundings that is constructed and/or controlled by humans. These elements have an expected behavior and are predictable, conditional on proper function. Examples of elements in this section include road infrastructures, the state of the road surface, and flying objects.

Each of these items may be seen and are to be interacted with within the built environment. The traffic infrastructure contains information pertaining to the road and traffic ahead making them valuable safety assets. The road surface is the most common element within most expected operational design domains (ODDs) for automated vehicles. Any element appearing on the road and the condition of the surface itself should be noticed by an AV to optimally navigate. Flying objects are elements found to be airborne, whether suspended in mid-air or moving with a trajectory. A descriptor of each category follows. 

\begin{figure}
    \centering
    \includegraphics[width=0.45\textwidth]{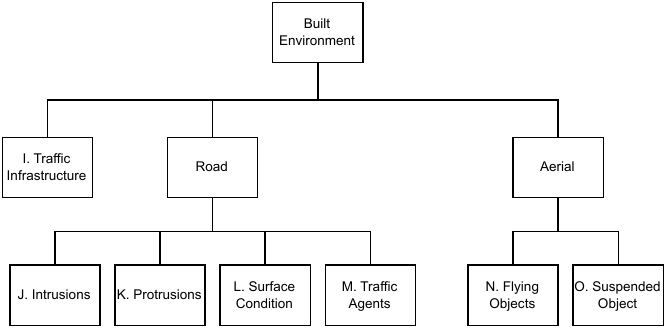}
    \caption{Taxonomy of elements occurring in the built environment.}
    \label{fig:tax_infrastructure}
\end{figure}

\begin{itemize}
\item[I.]{Traffic Infrastructure}
\end{itemize}

Traffic Infrastructure encompasses elements that are for traffic management and navigation. Infrastructure is used to guide drivers through traffic, increase the flow of traffic, and reduce accidents. The infrastructure consists of lines, curbs, traffic lights, traffic signs, and other ``2D'' and ``3D'' elements. It is important for an Autonomous vehicle to treat infrastructure as a source of information to better understand the context of the roads ahead in addition to (in many cases) objects to avoid \cite{crashstats_infr}.

The Manual on Uniform Traffic Control Devices for Streets and Highways (MUTCD) \cite{MUTCD} generally divides the traffic infrastructure into two parts: 1) signs and 2) markings. We will discuss how the two factors may play an adversarial role with regard to autonomous vehicle operations and what solutions have been proposed.

Street markings may vary across municipalities. Autonomous vehicles therefore require additional training before deployment into new areas. A more difficult problem is inconsistent, improper, or nonexistent road marking. Autonomous vehicles have been shown to accurately comprehend correct lane markings and road boundaries in real-time using their perception systems \cite{Huang_Lidarlane, kaddis2022developing, rao2022developing, shah2023comparing}, even if the markings are contradictory \cite{Kim_lanedetect} or nonexistent \cite{Assidig_Cameralane}.

Signs and signals are used for safe navigation and traffic flow management. Understanding and properly following this information is important, and perception systems have been used to solve this problem for some time\cite{Escalera_sign}, with modern improvements using machine learning techniques \cite{Wali_sign}. Like markings, signs can have irregularities as well such as a state disrepair, vandalism, or weathering \cite{Magnussen_sign}. \cite{Eykholt_sign} determines how little changes to signs create AV-hostile infrastructure. On the other hand, \cite{Lu_sign} demonstrates that this problem could be confined to recognition in images rather than video where an object would be seen in a continuous series of angles and distances. Therefore, it is possible autonomous vehicles will not be misled by altered signs and signals.

To address challenges resulting from a lack of communication between and among vehicles, infrastructure, and pedestrians, one solution is V2X connectivity \cite{Siegel_its_survey}. This solution comprises in-car connectivity units (``On Board Units [OBUs]''), and roadside units (RSUs), and may rely on individuals' existing mobile and other computing devices to increase the sensing and perception horizon and reduce latency and ambiguity in messaging, e.g. by reporting moving or stationary hazards through dense fog or at night via a standard communication protocol, semantic technology~\cite{mayer2015conversations}, or platform~\cite{siegel2013cloudthink, wilhelm2015cloudthink, siegel2016data, siegel2011design, siegel2018algorithms}. Other applications can be used to visualize the state of a vehicle and its environment, including hazards, remotely~\cite{pappas2018virtualcar}. These data can be valuable, but also sensitive in nature~\cite{soley2018value}.

\paragraph*{4.2.2.1 Road}

Road as a part of built infrastructure comprises all elements that an autonomous vehicle interacts with on the pavement. This can include surface conditions, debris, and other vehicles. We have divided this class into 4 elements: Intrusions, Protrusions, Surface Condition, and Traffic Agents, detailed in the following passages.

\begin{itemize}
\item[J.]{Intrusions}
\end{itemize}

Intrusions are road inconsistencies formed by erosion, wear, or damage on the surface of the road producing cavities such as potholes. Some cavities are formed from normal wear and tear, and are expedited in cold conditions when water seeps into cracks, expands when frozen, and forces the surface apart. A driver can decide to avoid the pothole and prevent possible damage to the vehicle \cite{bbc_pothole} and increase comfort. 

Some manufacturers use active suspension systems to better absorb the forces of falling into and climbing out of potholes; some sensing and perception systems can be designed to detect, alert, or avoid these road surface intrusions \cite{eriksson2008pothole, Dewangan_potnet, Talha_pothole}. \cite{Madli_pothole} demonstrates how using an ultrasonic sensor potholes can be detected and the driver be warned to take evasive action. Using a GPS sensor they also record geological data to remember previously encountered potholes. This information can be shared between vehicles and vehicles can cooperate to avoid potholes as a group\cite{Siegel_its_survey}.

\begin{itemize}
\item[K.]{Protrusions}
\end{itemize}

A protrusion is a road inconsistency that sits above the expected average surface height of the road. Planned protrusions include speed bumps, curbs, and rumble strips, while unplanned protrusions can be debris or extruding portions of the road surface itself. Detecting planned protrusions could be easy as they are often accompanied by signs and markings that warn of their existence. Unplanned protrusions are difficult to detect, as they could have a random size and shape or be mobile. 

Autonomous vehicles will face protruding obstacles and should be able to safely navigate around them as well. \cite{Yu_protrusion} shows how elements like ambient light, weather, and sensor accuracy can affect the performance of the vehicle in obstacle avoidance. Despite these difficulties, research has shown using segmentation \cite{Islan_protrusion} and sensor fusion \cite{Yu_protrusion} can improve this performance even in real-time scenarios \cite{Abbas_protrusion}. Connectivity can also be used to share details of protrusions and other hazards beyond a vehicle's field of sight \cite{khalfin2023vehicle}. 

Training vehicles on scenarios with variations of planned and unplanned protrusions becomes important as we aim for level 5 autonomous driving systems \cite{sae}. Simulating these scenarios can help with this objective as demonstrated in \cite{Abbas_protrusion} and can prove autonomous vehicle's safety potential before they are deployed.

\begin{itemize}
\item[L.]{Surface Condition}
\end{itemize}

Surface condition is concerned with the material (asphalt, concrete, dirt), quality \cite{Paser}, and condition (dry, wet, icy, snowy) of the traction surface. These aspects of the road surface are important to consider as each condition could require different planning and control with respect to the road traction \cite{crashstats_infr}.

Systems have been developed for mitigating loss of traction like Traction Control (TC), Electronic Stability Control (ESC), and Anti-Lock Braking System (ABS). These systems actively monitor the stability and grip of each wheel, and in cases, apply control inputs to regain traction. Such systems do not take full control of the vehicle and will not necessarily bring the vehicle back to a safe state without external input.

Autonomous systems however have direct command over the vehicle inputs and have access to stability and traction data. Thus, can perform better than humans under uncertain road surface conditions. Research has shown how autonomous vehicles can be trained to deal with limited traction on road surfaces \cite{Berntop_traction}. \cite{Zhu_traction} also demonstrates autonomous vehicle off-roading capabilities using multiple deep-learning models.

\begin{itemize}
\item[M.]{Traffic Agents}
\end{itemize}

Even though signs and road markings attempt to control the flow of traffic, traffic agents may be confused or choose to not follow them. This can turn into a hazard for other road users. Autonomous vehicles have to monitor all surrounding vehicles not only to plan their own trajectory but to measure the traffic's trajectory and plan to avoid them, assuming the worst~\cite{9991836}. 

V2X solutions can alleviate this concern when surrounded by other connected cars. Autonomous vehicles can communicate with other vehicles, or with infrastructure, gather environmental data, and plan their controls with much more accuracy than humans might be able to \cite{Siegel_its_survey}. Connected vehicles can also plan their movement cooperatively and result in a more efficient performance like platooning on highways \cite{Olia_platooning}.

Special cases like emergency vehicles require a different response such as yielding or pulling over no matter the circumstance. Failure to recognize, and safely yield to emergency vehicles could have dire consequences as there may be people on the street \cite{Tesla_crashes}. This makes it important to recognize first-responder vehicles by their lights or sirens \cite{Sun_emergency}.

\paragraph*{4.2.2.2 Aerial}

Aerial elements ``float'' in the air. It not only contains the usual elements such as drones, bridges, and parking gates but also could contain elements that may not be normally expected, such as a detached wheel bouncing on the road or airborne debris. This is further categorized as flying or suspension.

\begin{itemize}
\item[N.]{Flying Objects}
\end{itemize}

Flying objects may have any origin, size, or speed. A flying object could be very dangerous especially if it makes an impact with a vehicle's windshield. As long as flying objects have an impending trajectory, an autonomous vehicle may have to take evasive action. For example, traffic coming to a stop for a plane making an emergency landing on the road \cite{Plane_onRoad}.

The majority of the work done on detecting and identifying flying objects has been done for unmanned aerial objects (UAVs). This is either done for the detection of UAVs and/or for detection by UAVs. However, the technological mechanism of vision and processing is similar between UAVs and autonomous vehicles making advancements in these two fields relatable. This is demonstrated as UAVs have been shown to be detectable as flying objects using a camera \cite{Rozantsev_fly_cam} or LiDAR \cite{Vrba_fly_lidar} onboard other UAVs. This eases adding airborne object detection abilities on autonomous vehicles.

\begin{itemize}
\item[O.]{Suspended Objects}
\end{itemize}

Suspended objects are any objects above ground that are still and are expected to stay still unless interacted with. These objects can be obstacles like parking gates or high structures like hanging cables and overpasses \cite{Truck_vs_Bridge}. Low-hanging obstacles are objects that an AV would have to detect and avoid. Whereas the height of higher objects would have to be measured by the AV to determine the clearance underneath.

A truck's tail lift is a difficult obstacle to identify since it may be suspended at any height. Moreover, the relative abundance of delivery trucks with tail lifts in urban makes them a likely element to encounter. To this end \cite{Sun_elevation} developed a radar system with improved accuracy in estimating obstacle elevation. This allows the autonomous system to better approximate the safety of the ``drive-under'' maneuver required for objects suspended above ground.

The built environment contains some elements that the AV would have the most interactions with. Failure of recognition and execution of proper responses to these elements could undermine the safety of lives and properties. The use of simulations to recreate the built environment and its elements can fast-track the training process of AVs since fringe examples are hard to come by naturally.

In this section, we introduced a taxonomy of adversarial elements that autonomous vehicles may encounter. This taxonomy is intended to be highly inclusive. While the set of ``unknown-unknown'' adversarial events is infinite, this approach appears to encapsulate many ``known'' events and those yet unseen. To the best of the author's knowledge, this taxonomy is the best example of such a format to dissect events into their smallest elements.

The 15 element classes mentioned are categorized into three parts connected to a hierarchically shared parent to construct the taxonomy. Each class of adversarial elements is explained along with current standards and/or state-of-the-art research in threat mitigation threats. The importance of recreating scenarios with these elements has been highlighted and an example of how said elements can become adversarial is provided.

This taxonomy is designed to be used by the developers of autonomous vehicles to ensure the safe operation of their vehicles in any scenario. This taxonomy could be used as a consistent framework to analyze and discuss complex and adversarial events as their contributing elements, easing the process of understanding why and how much each existing element has a responsibility in turning a normal event into an adversarial one. In an alternate use case, the taxonomy can be used to envision and recreate adverse scenarios in simulation and reality for training, testing, and evaluating an autonomous system's abilities. 

\section{Discussion}

The primary purpose of this taxonomy is to help identify which types of adversarial events an AV needs more training against. AVs can safely traverse the roads in predictable scenarios and operating contexts, while anomalies, however simple, might be better handled by a human driver without advanced AV training. It would be beneficial to be able to identify and generate these anomalous cases using a taxonomy, as such cases can be recreated in simulation \textendash thereby allowing AV systems to reinforce their understanding of such events and discover and apply safe measures. The efficacy of such taxonomy in identifying all types of anomalous events at a large scale would validate its usefulness. An experiment has been conducted accordingly and results presented. In this section, we will discuss the implications and limitations of the taxonomy, conduct an evaluation of the taxonomy, and interpret the results of the evaluation.

\subsection*{Taxonomy}
This taxonomy can serve as a guide for AV companies to identify when their vehicles lack exposure to certain events from limited training data. This taxonomy is meant to help the developers build an understanding of AV system misbehavior and shortcomings either on the road towards other agents and objects, towards its passengers' benefits and well-being, or in the way it conducts itself with respect to the environment. Provided in the appendix are two accidents involving autonomous vehicles, which are examined, and their contributing elements are identified using the taxonomy. A similar process can be repeated by AV developers using established adversarial events, recreating similar events in simulations by combining elements of the taxonomy to generative diverse training data including underrepresented scenarios, reducing their vehicle's failure rate.

Each leaf of the taxonomy is paired with examples of contemporary academic research or applicable standards on the topic. The authors do not stipulate any solution of their own. The provided studies for each element are not presented as a perfect and aggregate solution to any problem caused by those elements. There might exist unmentioned safe approaches for how to safely conduct self-driving in the face of event's diverse variations. The presented studies are rather representative of the possible danger of said elements and the work being done toward their understanding and mitigating strategies. Therefore finding an adequate and safe solution for each adverse event is left at the discretion of whoever uses this taxonomy in the development of their product.

It is important to note some of the classes may happen in parallel such as adverse weather, reduction of ambient lighting, and change in the condition of the road. However, these classes were carefully selected not to be mutually inclusive. An example would be encountering a wet road surface but no weather conditions.

In the natural events category animals are excluded and separated from VRUs because VRUs have the cognitive capacity to identify a vehicle, consider the possible consequences, and make an effort to avoid a collision if they choose to. Animals may not have such a level of contextual awareness. Furthermore motorcycles and included in VRUs instead of traffic agents as they have an increased vulnerability compared to 4-wheeled vehicles.

Although traffic agents are either controlled by human operators or autonomously they are not categorized under natural environments like VRUs. This is because VRUs are often smaller, have a higher degree of mobility, and can tread on more varied areas where larger vehicles may not be able to. The inferior mobility of vehicles reduces the set of possible random mistakes by their drivers as their operational design domain is more often restricted to built infrastructure. Furthermore, traffic agent drivers are also not considered as a separate entity from their vehicles. This is because the actions of the traffic agent driver are displayed through their vehicle. Therefore, following the patterns displayed by the traffic agent should describe the intention of their drivers too.

Suspended objects may at times move and turn into flying objects such as parking barrier gate arms. This means an object can change element class at a moment's notice. A suspended object may also fall as a result of a breakdown. Such an object would turn into a flying object in mid-air and subsequently turn into road debris (protrusion) once on the road surface. This exhibits how the classes are designed to be unique but they may still exhibit some correlations.

User modifications to the vehicle are not directly considered. Autonomous vehicles are accompanied by user guidelines and the users are made to acknowledge the limitations of these vehicles. Modification of an autonomous vehicle may alter its functions and cast doubt on its safe operation. Thus it might mean void warranty and disable complex capabilities (such as ADS) of the vehicle if the modifications are unauthorized. This was covered in \cite{Capallera}. These modifications can be interpreted to be covered by maintenance, software, or mechanical health.

This taxonomy attempts to categorize all elements outside of the "unknown-unknown" set of adversarial elements. There are no limitations to "unknown-unknown" driving scenarios; scenarios that not only not normally expected but also have never been identified. Thus this taxonomy might fail to identify a newly discovered "unknown-unknown" adversarial element as seen in the results. \cite{Ramirez} recognizes these events as ``Unpredictable Environment''. We, on the other hand, do not incorporate an \emph{unpredictable} class in our taxonomy since we want to build an exhaustive taxonomy. Since the newly found ``unknown-unknown'' elements may be random and unorganized, an unpredictable class would not be helpful to developers of autonomous vehicles. Consequently, we encourage the users of this taxonomy to make alterations to the taxonomy and add any newly found ``unknown-unknown'' element with respect to their operational design domain.

\subsection*{Evaluation}
In this section, we assess how well the presented taxonomy can be used to identify scenarios where multiple adversarial elements may appear. The difficulty of identifying the most significant element in a scenario implies the irregularity of the event.

The data used to evaluate the taxonomy is from the California DMV autonomous vehicle collision reports dataset \cite{CaliforniaAVReport} from January to September of 2023. The selected data was held out from the data used in training to serve as validation data. The dataset had 117 samples of collisions involving autonomous vehicles. One sample had demonstrated possible mistakes and inconsistencies in report details and was thus removed. The remaining 116 samples are used to analyze the practicality and legitimacy of the taxonomy. A link to the evaluation dataset can be found in the appendix.

We classify a scenario's contributing elements according to the proposed taxonomy. The difficulty of classification due to multiple existing elements or vagueness of the elements in the scenario is communicated using a subjective numerical complexity factor shown in table \ref{tab:difficulty}. Higher numbers express higher complication for fit, with ``4'' representing scenarios where an element could not be reliably and repeatably classified by the taxonomy. The difficulty scale is subjective to the lead author's interpretation of the sample events and its ability to fit into the taxonomy as presented, without modification or ambiguity. Each entity is encouraged to develop a complexity scale corresponding to their interpretations and their product's capabilities. Also in this table, we can see the number of samples corresponding to each difficulty level.

In table \ref{tab:classes} we find the the occurrence of each class. This is based on the number of samples whose most significant contributing element is found in the taxonomy.

\begin{table}
\begin{center}
\caption{Difficulty level of sample classification and corresponding sample numbers}
\label{tab:difficulty}
\begin{tabular}{|c|c|c|}
\hline
\textbf{Difficulty} & \textbf{Numerical Grade} & \textbf{Number of Samples}\\
\hline
Easy & 1& 95\\
Moderate & 2 & 18\\
Difficult & 3 & 1\\
Indecisive & 4 & 2\\
\hline
\end{tabular}
\end{center}
\end{table}

15 samples only had one element present in their scenario. The other 101 samples had multiple possibly contributing elements present at the scene of the accident. This exhibits how the scene of an accident is often not simple. Rather a mixture of multiple elements appearing in series or parallel with various characteristics could contribute to an accident.

\begin{table}[b]
\begin{center}
\caption{Number of samples classified by each class ordered by class appearance in taxonomy}
\label{tab:classes}
\begin{tabular}{|l|c|}
\hline
\multicolumn{1}{|c|}{\textbf{Class of Element}} & \textbf{Number of Samples}\\
\hline
B. Software & 6\\
D. Driver & 14\\
G. VRUs & 10\\
H. Animal & 1\\
I. Traffic Infrastructure & 1\\
J. Protrusion & 4\\
K. Intrusion & 2\\
M. Traffic Agent & 72\\
N. Flying Object & 2\\
O. Suspended Object & 2\\
\textbf{Unclassified} & \textbf{2}\\
\hline
\end{tabular}
\end{center}
\end{table}

Results show two samples had elements that were not identified in the taxonomy. Both these samples were instances of a parked vehicle's door being opened into the immediate trajectory of the ego vehicle. This left the ego vehicle with not enough time to react and it made contact with the parked vehicle's opened door. When a passenger of the vehicle opens the door, it becomes difficult to easily interpret this case as either VRU or a traffic agent. In both cases, the vehicle blurred the line between external agents and other elements of the natural environment, suggesting a potential point of ambiguity in the taxonomy. Here, the door itself may be interpreted as a suspended object or an extension of the parked vehicle. This can be potentially classified as an unpredictable event as it is in \cite{Ramirez}, though our taxonomy lacks such a class. 

As these events were not included in the initial revision of the taxonomy but were present in the hold-out validation incidents, we propose to expand the definition of \emph{Traffic Agents} to include parked or inactive traffic agents and all extensions to the traffic agents such as doors, trailers, and attachments. While this alteration addresses the two scenarios not characterized by the taxonomy's initial version, it is possible and likely that additional scenarios will emerge with difficult-to-characterize contributing factors. This is a reflection on the complexity of the problem space, and an important reminder for users of this and other taxonomies and scenario generation tools to be thoughtful, exhaustive, and adaptive to ensure that development exposes learning systems to even those training and test cases yet unimaginable. 

Assuming this change was made to the definition of Traffic Agents within the taxonomy, the previously unclassified samples would then be classified as traffic agents. This is not reflected in the dataset as posted online, as we wished to share the original validation set without biases introduced through the lens of the revisions the outlying events' characterization inspired. This maintains the integrity of the initial classification evaluation result. 

\section{Conclusion}
In this paper, we presented a taxonomy for organizing adversarial events that can provide this framework, by grouping similar events and providing a consistent format to discuss and analyze them. The taxonomy organizes the contributing elements of an event into 15 different classes among three categories: Ego, Natural Environment, and Built Environment. We also showcase the benefits of using a taxonomy, such as enabling the grouping of similar events and the ability to uniformly atomize an event into its contributing elements. Thus making it simpler to imagine complex scenarios involving one or more adversarial events occurring in parallel or series, identifying patterns, commonalities, and potential vulnerabilities across different types of events, prioritizing and focusing efforts on the most critical risks, and identifying potential areas for improvement.

We demonstrated how this taxonomy can be used to dissect contributing elements of adverse traffic scenarios and generate new training data for critical safety algorithms using simulations. Doing so will provide a more comprehensive view of the risks and challenges that automated vehicles may face and will make it easier to simulate and test variations of these complex scenarios, improving the safety and reliability of the automated vehicle's system. To display the merit of the taxonomy we test the taxonomy against 116 samples of autonomous vehicle accidents. We obtained a 98.3\% rate of success in detecting the elements identified within sample hazard events based on the initial version of the taxonomy. Hence, we expanded the taxonomy to reduce ambiguity in the case of the two unseen contributing factors. 

Our research provides a taxonomy for organizing adversarial events in the context of automated vehicle development and highlights the importance of understanding and anticipating different types of adversarial events to design systems that are robust and resilient to them. This can help to ensure that automated vehicles are safer and more reliable for the public. However, there are many directions for future research that can be taken to improve the taxonomy and its use in automated vehicle development.

\section{Future Work}
One potential direction for future research is incorporating machine learning techniques to improve the classification of adversarial events. This can help to automate the process of classifying events and make it more accurate and efficient. Additionally, the taxonomy could be expanded to include additional classifications, such as the level of control an attacker may have or the likelihood of an event to occur.

Another area of future research is to evaluate the taxonomy's performance in different applications and environments, such as in motorsport settings or off-road. This can help to understand how well the taxonomy performs in different scenarios and identify any areas for improvement.

It is important to note that the research has potential implications for the broader field of automated system development, specifically in terms of the safety and reliability of the systems. The use of taxonomy can help developers anticipate and prepare for different types of adversarial events, and simulating and testing these complex scenarios can improve the safety and reliability of broader automated systems.

\section*{Acknowledgements}
We thank the present and past members of the MSU DeepTech Lab and Trusted Systems Lab for their assistance in brainstorming and feedback on the creation of the taxonomy.

\section*{Appendix}

\subsection*{Evaluation Data Reference}
The dataset used for the evaluation of the taxonomy is saved as a CSV file and hosted on Harvard Dataverse \cite{Eval_Dataset}. Alongside the dataset, a text file is provided with information about the data origin and use.

\subsection*{Using the Taxonomy to Generate Training Data from Scenarios}
Here, we discuss two example scenarios that demonstrate the use of our taxonomy for organizing existing adversarial events and in generating training data for critical safety algorithms. These scenarios are widely publicized accidents chosen for their variety and difficulty in identifying fundamental contributing elements. We explain how the taxonomy was used to identify and classify the environmental elements in each scenario and how the data from these scenarios can be used to train critical safety algorithms.

The elements found in each example are \textit{italicized} for emphasis. Both incidents are reported based on the best available knowledge. There is no judgment or blame implied in the description of what was reported and believed to have occurred in each incident. 

\subsubsection*{Cruise, October 2nd 2023}
On the night of October 2nd, 2023, an autonomous vehicle developed by Cruise LLC carrying no passengers was involved in an accident with a pedestrian \cite{cruiseAccident}. A jaywalking pedestrian proceeding through an intersection after the light turned green and was hit by a vehicle to the left of the AV. The pedestrian was thrown into the path of the AV, which detected the pedestrian and applied the brakes but made contact. After coming to a stop, the AV attempted to pull over to the curb while the pedestrian was pinned under the vehicle, dragging the injured pedestrian for 20 feet.

In this scenario, multiple elements may have contributed directly and indirectly. The \textit{pedestrian} element was central, and they had the right of way. The \textit{vehicle} next to the AV made the first contact with the pedestrian. Although the AV can not control other traffic agents' behavior, it can predict their trajectory and often sense impending contact. Although the \textit{autonomous system} detected the pedestrian to be in its path, applied the brakes, and came to a stop, it remains unclear whether it recognized the pedestrian to be stuck under the vehicles before proceeding forward. The \textit{road infrastructure} was present at the scene, as the vehicles were stopped before the traffic lights turned green. Finally, as this accident happened at night there were no natural ambient \textit{lighting} and visibility was dependent on street lights and vehicle lights.

The taxonomy can be used to recognize the aforementioned five elements present at the scene of an accident belonging to these classes: VRUs, Traffic Agents, Software, Traffic Infrastructure, and Ambient Lighting. Each element can be isolated and further explored while training using alternative versions of this scenario. The characteristics and behavior of each element can have variations in simulation to produce different data from similar scenarios. For example, a pedestrian that does not jaywalk, or is initially in the path of the AV instead of the adjacent vehicle, or manages to dodge the adjacent vehicle. Various scenes from later in the scenario can be recreated too like having debris pinned under the vehicle instead of a pedestrian or having the pedestrian being thrown onto the vehicle instead of being pinned under. 

\subsubsection*{Tesla, March 1st 2019}
At dawn of the morning on March 1st, 2019, A Tesla driving in autopilot mode on U.S. Route 441 made contact with a truck and trailer causing the death of the Tesla driver, Jeremy Banner \cite{Banner_crash}. Banner had turned on the L2 automation system after setting the speed to 69 mph on a road with a 55 mph speed limit. The specific version of the Autopilot equipped on the vehicle was not designed to drive on that type of roadway, though it was allowed to be engaged. The vehicle detected Banner not holding the steering wheel. Meanwhile, a truck with a trailer in cross-traffic attempts to make a left turn into the oncoming lane of the Tesla without stopping at a stop sign or yielding to traffic. Autopilot failed to correctly recognize the truck as an impending object on the road. With neither the Autopilot nor Banner applying the brakes, the Tesla hits the trailer, passing underneath and resulting in death for the Tesla occupant. 

The \textit{Tesla driver} had engaged the Tesla Autopilot but had shown not to be paying attention to the road ahead. The vehicle is an SAE level 2 \cite{sae} vehicle and requires the driver to be attentive. The truck driver also paid limited attention to the oncoming traffic and the \textit{stop sign}, thus making the \textit{truck} a dangerous element to the other road users. The \textit{Autopilot system} was not well suited to the operating context and stayed engaged while driving at 14 mph above the speed limit. Furthermore, the autonomous system failed to detect the truck on the road and apply the brakes. Since this accident happened at dawn, natural \textit{lighting} is very little thus the only source of light would be the vehicles' headlights.

The taxonomy can identify elements from five classes in this scenario: Driver, Traffic Agents, Traffic Infrastructure, Software, and Ambient Lighting. Variations of scenarios can be used to train autonomous systems in a simulation like having attentive drivers or a setting with brighter ambient lighting. Alternative traffic infrastructures could be tested too such as traffic lights instead of stop signs or if the scenario were to occur in a city setting with a lower speed limit.

The two examples above are identified to have similar classes of adversarial elements in their scenarios. Nevertheless, they each are different examples of each class. This demonstrates how a class of elements can appear in many different ways and would require different types of responses, necessitating unique training approaches.

\Urlmuskip=0mu plus 1mu
\bibliographystyle{IEEEtran}
\bibliography{bibliography.bib}

\vfill

\end{document}